%% file: main.tex
\documentclass[conference]{IEEEtran}
\IEEEoverridecommandlockouts

\usepackage{cite}
\usepackage{amsmath,amssymb,amsfonts}
\usepackage{algorithmic}
\usepackage{graphicx}
\usepackage{textcomp}
\usepackage{xcolor}
\usepackage{helvet}
\usepackage{courier}
\usepackage{tabularx}
\usepackage{booktabs} 
\usepackage{subfig}
\usepackage{float}
\usepackage[square,numbers,sort&compress]{natbib}

\newcommand{\Graph}{\mathcal{G}}
\newcommand{\Vertices}{\mathcal{V}}
\newcommand{\Edges}{\mathcal{E}}

\def\BibTeX{{\rm B\kern-.05em{\sc i\kern-.025em b}\kern-.08em
    T\kern-.1667em\lower.7ex\hbox{E}\kern-.125emX}}

\begin{document}
\title{Learning traffic flows: Graph Neural Networks for Metamodelling Traffic Assignment}

\author{%
  \IEEEauthorblockN{Oskar Bohn Lassen\IEEEauthorrefmark{2}*, 
  Serio Agriesti\IEEEauthorrefmark{2}*, 
  Mohamed Eldafrawi\IEEEauthorrefmark{3}, 
  Daniele Gammelli\IEEEauthorrefmark{4},\\ 
  Guido Cantelmo\IEEEauthorrefmark{2}, 
  Guido Gentile\IEEEauthorrefmark{3}, 
  Francisco Camara Pereira\IEEEauthorrefmark{2}}
\\
  \IEEEauthorblockA{%
    \IEEEauthorrefmark{2}Department of Technology, Management and Economics,\\ Technical University of Denmark, Lyngby, Denmark\\[0.5ex]
    \IEEEauthorrefmark{3}Department of Civil, Constructional and Environmental Engineering,\\ Sapienza University of Rome, Rome, Italy\\[0.5ex]
    \IEEEauthorrefmark{4}Department of Aeronautics and Astronautics,\\ Stanford University, Stanford, California}
}

\IEEEoverridecommandlockouts
\maketitle
\IEEEpubidadjcol

\begin{abstract}
The Traffic Assignment Problem is a fundamental, yet computationally expensive, task in transportation modeling, especially for large-scale networks. Traditional methods require iterative simulations to reach equilibrium, making real-time or large-scale scenario analysis challenging. In this paper, we propose a learning-based approach using Message-Passing Neural Networks as a metamodel to approximate the equilibrium flow of the Stochastic User Equilibrium assignment. Our model is designed to mimic the algorithmic structure used in conventional traffic simulators allowing it to better capture the underlying process rather than just the data. We benchmark it against other conventional deep learning techniques and evaluate the model's robustness by testing its ability to predict traffic flows on input data outside the domain on which it was trained. This approach offers a promising solution for accelerating out-of-distribution scenario assessments, reducing computational costs in large-scale transportation planning, and enabling real-time decision-making.
\end{abstract}


\section{Introduction}
A central problem in traffic management is the Traffic Assignment Problem (TAP). TAP deals with determining the distribution of traffic flow in a transportation network that meets a specific equilibrium condition \citep{sheffi1985urban}. Many of the current solutions struggle with scalability and adaptability to real-world dynamic traffic conditions \citep{fan2022large} and considerable work has been devoted to identifying effective algorithms for this issue, as highlighted in \citep{gentile2014local,babazadeh2020reduced}. Variations of the TAP are the stochastic user equilibrium \citep{damberg1996algorithm,gentile2018new}, multi-modal traffic assignment \citep{pi2019general}, and dynamic traffic assignment \citep{gentile2007spillback,ben2012dynamic,gentile2010general}. These adaptations aim to integrate factors such as variability in travel times and driver behavior offering traffic assignment solutions that are closer to real-world conditions. 
In this context, the high availability of traffic data \citep{wu2018hierarchical,kostic2015using,mitra2020methodology,ros2022practical} has recently fostered the emergence of data-driven techniques to address various challenges in transportation. As the machine learning field advanced, deep learning models \citep{veres2019deep} such as multi-layer perceptron (MLP) and Recurrent Neural Networks (RNN), including specialized forms like Long Short-Term Memory (LSTM) and Gated Recurrent Units (GRU) \citep{sharma2023GRU, Waikhom2023forecasting}, have been developed. These models, while effective in handling temporal sequences, struggle with spatial correlation. To bridge this gap, \citep{ye2019co} introduced an approach that combines Convolutional Neural Networks (CNN) with LSTM. However, transportation networks are inherently graph-structured, with roads, intersections, and travel demand forming a complex relational system. Graph Neural Networks (GNNs) have emerged as a pivotal technology in transportation research, as they natively operate on graph-structured data, making them well-suited for problems such as traffic assignment, network flow estimation, and transportation planning \citep{Rahmani2023lit,Jiang2023lit}. 
GNNs have been utilized in different applications in the transportation domain \citep{veres2019deep}. Some examples are: \citep{Ye2023accidents} for accidents, speed \citep{Waikhom2023forecasting}, flow \citep{Chen2023nodes, Huang2023multiday} and density \citep{Khan2023GGNN} forecasting, \citep{Gammelli2021AMoD} for automated mobility on-demand control problems, and \citep{Ouyang2023cities} for cross-city knowledge transfer. While the application of GNN to TAP problems is still barely addressed in literature, some studies stand out.
Specifically relevant for this work is \cite{rahman2023data}, which exploits a graph convolutional neural network (GCN) to estimate the traffic flow in a simulation setting similar to the one adopted in this paper. Still, as it will be showed, the aim of the two studies differ as the authors of \cite{rahman2023data} aim to solve a prediction problem in-distribution while we assess the performance of machine learning models out-of-distribution and introduce an architecture that is specifically designed to improve the out-of-distribution performance. Finally, one key difference is that, while \cite{rahman2023data} considers each junction as a centroid (i.e. a specific ID entry in the OD matrix), we treat junctions and centroids separately which more closely resembles standard practice in transportation (indeed, rarely OD data is available at junction level). In \cite{liu2024capacity}, instead, an heterogeneous GNN is designed to consider virtual links as edges to pass the information between nodes in OD pairs. The study is relevant as it incorporates capacity within the GNN architecture in a similar fashion as is proposed in this work. Still, in the presented work speed limits are added as edge features and the hypothesis of each junction as a zone in the OD is relaxed. Finally and most importantly, \cite{rahman2023data, liu2024capacity} both employ a different algorithm for traffic assignment (Frank-Wolfe), while the proposed GNN learns to approximate the SUE.
Another relevant mention is \cite{jungel2024wardropnet} which combines neural network techniques with a combinatorial optimization layer that uses the estimation of flow from the neural network layers to solve an equilibrium problem, thus incorporating the search for equilibrium within a neural network architecture. Few other works attempt to address TAP using machine learning approaches such as MLP \citep{su2021deep} and CNN \citep{fan2023deep}, but these models do not efficiently leverage the topological information and relational features among nodes and edges in the network, which are crucial for accurately modeling traffic dynamics (as mentioned, for example, in \citep{Wang2023topological, Chen2023nodes}). Additionally, many of the models in literature are limited to fixed network topologies, requiring retraining with any alterations in the network's configuration (e.g. a wide roadwork). As a result, these models may fail out-of-distribution as they do not fully encapsulate the complexity of traffic dynamics, which would instead by captured through a model able to replicate the TAP. \\ The presented work achieves the following. First, we leverage the power of the proposed Message-Passing Neural Network (MPNN) architecture to replicate the inner workings of a traffic simulator's TAP algorithm. Doing so allows the design of a GNN targeted towards generalizing to out-of-distribution cases, e.g. for variation in key quantities such as demand, speed limits and link capacities. Second, we incorporate edge features (i.e. free-flow travel time, capacity, and speed) into the MPNN model and assess how doing so impacts out-of-distribution performance. Third, we abandon the common design hypothesis found in literature that each junction in the network is a zone (i.e. an originator or attractor of demand). Indeed, this hypothesis is not true in most simulation experiments, especially for medium to large networks. Finally, we test the effects of the described features by comparing the quality of the results from the MPNN with other ML baseline models, in- and out-of-distribution. 

\section{Notation and theoretical background}

\subsection{The Traffic Assignment Problem}
The TAP plays a crucial role in determining how traffic volumes or flows are distributed across edges within a network. Consider a transportation network represented as a graph $\Graph = (\Vertices, \Edges)$, where nodes $\Vertices$ denote intersections or junctions of roads, and edges $\Edges$ denote the roads or links connecting these points. The objective of traffic assignment is to optimally allocate traffic flows to these edges. Different formulations of the TAP can be tailored to specific objectives and constraints. In the context of Stochastic User Equilibrium (SUE), the formulation takes into account the uncertainty and variability in drivers' perceptions of travel times and decisions. The SUE framework formulates an optimization problem to capture the probabilistic nature of travel time perceptions among drivers \citep{sheffi1985urban}. The goal is to distribute traffic flows such that no driver can unilaterally reduce their expected travel cost by choosing an alternative route. This condition is referred to as the Wardrop equilibrium. Mathematically, the SUE can be expressed as:
\begin{equation}
    \min_{\{f_e\}_{e \in E}} \sum_{e \in E} \left( \int_{0}^{f_e} t_e(w) dw + Var(f_e) \right)
\end{equation}
\begin{align*}
    &\text{subject to:} \\
    &\sum_{k} f_{rs}^k = q_{rs}, \hspace{0.4cm} f_{rs}^k \geq 0, \hspace{0.4cm} f_e = \sum_{rs} \sum_{k} f_{rs}^k \cdot \zeta_{rs}^{e,k}\\
    &\hspace{2cm} \forall k, r, s \in \Vertices, \hspace{0.2cm}\forall e \in E
\end{align*}

Here, $t_e(f_e)$ is the travel time function for edge $e$, $Var(f_e)$ represents a term that captures the variance or uncertainty in travel times (namely how different the perceived travel time is from the actual one), $q_{rs}$ denotes the demand from source $r$ to destination $s$, $f_{rs}^k$ indicates the flow on the $k^{th}$ path from $r$ to $s$, and $\zeta_{rs}^{e,k}$ is a binary indicator equal to 1 if edge $e$ is on the $k^{th}$ path connecting $r$ and $s$.

\subsection{Graph Neural Networks}

In this study, we argue that GNNs are particularly well-suited in the context of traffic assignment, as they excel by mimicking the process of traffic flow distribution. They do this through iterative updates that simulate traffic moving from one node to another, adjusting the distribution based on the current state of the network and crucially the constraints on each link, such as capacity limitations and expected travel times. Given a graph \(\Graph = (\Vertices, \Edges)\), with \(\Vertices = \{v_i\}_{i=1}^{N_v}\) representing the nodes and \(\Edges = \{e_k\}_{k=1}^{N_e}\) the edges, GNN models aim to learn a permutation-invariant function. This function processes a \(D\)-dimensional feature vector \(x_i\) for each node, alongside a structural representation of the graph, typically through an adjacency matrix \(A\), to produce updated node representations \(x'_i\).

\section{Methodology}
This section outlines our proposed framework for predicting traffic flow arising from the SUE. We base the model on the SUE implemented in PTV VISUM, a widely used traffic simulator \cite{ptvvisum}. The specific architecture used in this work is the Message-Passing-Neural-Network (MPNN), a class of Graph Neural Networks (GNNs) that supports learning on graphs by directly integrating both node features and edge attributes into the learning process. 
The model architecture is building on the GraphGPS framework proposed by \cite{rampasek2022GPS} which is composed of three key components: an Encoder, an MPNN layer, and a Decoder, each playing an important role in processing and interpreting the network data. The MPNN layer executes a sequence of steps: message passing, aggregation, and update, each needed to capture and integrate complex relational patterns between nodes and edges. In the following, a more detailed analysis is provided.

\subsubsection{Input Representation} Given a transportation network modeled as a graph \(G = (\Vertices, \Edges)\),
for each node \(v_i \in \Vertices\), the feature vector \(\mathbf{x}_i\) captures the OD demand from node \(v_i\) 
to every other nodes in the network. A 'centroid node' represent the 'center of gravity' of a traffic analysis zone, 
and represents the beginning or end of individual trips in a specific zone. Centroid nodes are represented as nodes 
on the graph - similar to any other node - however, their feature vectors contain the outgoing OD demand to all 
other zones. Conversely, nodes that serve only a structural role in the network (i.e., non-centroid) are also included
as graph nodes in the model but their feature vectors are padded with zeros. This modeling decision reflects standard
practices in transportation modeling, where the demand is represented at the level of OD matrices between different
zones, and the centroids are the nodes that generate the demand to and from a certain zone. While this constraint introduces additional complexity during the model's training phase, it enhances the real-world applicability of the proposed MPNN, ensuring that the model aligns with practical data availability constraints. This approach offers an improvement over similar works in the literature, where such considerations are often neglected.

The feature vector for node \(v_i\) is a vector \(\mathbf{x}_i\) of length \(n\) (number of centroids and square root of the OD matrix's number of cells), where each element \(x_{ij}\) represents the demand from \(v_i\) to \(v_j\):
\begin{align}
\mathbf{x}_i = [x_{i1}, x_{i2}, \ldots, x_{in}]
\end{align}

The node feature matrix \(X\) is then an \(n \times n\) matrix with as many rows as origins and as many column as destinations. Namely, each vector \(\mathbf{x_{i}}\) is a row.
Edges \((v_i, v_j) \in \Edges\) include feature vectors \(\mathbf{e}_{ij}\) with the free-flow travel time \(T_{ij}\) representing the time taken to travel from node \(v_i\) to node \(v_j\) under optimal conditions, speed limit $S_{ij}$ representing the maximum speed to do so, and capacity \(C_{ij}\), indicating the maximum traffic volume that can traverse the link before congestion. The feature vector for an edge is thus defined as:
\begin{align}
\mathbf{e}_{ij} = [T_{ij}, S_{ij}, C_{ij}]
\end{align}

The adjacency matrix \(A\), a \(n \times n\) matrix, represents the connectivity between nodes, where an entry \(a_{ij}\) denotes the presence of an edge from node \(i\) to node \(j\), with \(a_{ij} = 1\) for unweighted graphs or \(a_{ij} = w_{ij}\) for weighted graphs indicating the weight of the edge. The adjacency matrix $A$ is not a node or edge feature in the usual sense but rather a structural representation of the graph itself. It encodes global connectivity and is used in GNNs to control where information can propagate between nodes during message passing (eq. \ref{eq:message}).

\subsubsection{Encoder} The encoder layer produces latent embeddings for node and edge features via a linear transformation, capturing relevant information for subsequent GNN operations:
\begin{align}
\mathbf{h}_i^{(0)} = W \mathbf{x}_i + \mathbf{b}
\end{align}

where \(\mathbf{x}_i\) represents the input feature vector of node \(i\), \(W\) is the weight matrix of the encoder layer, \(\mathbf{b}\) is the bias vector, and \(\mathbf{h}_i^{(0)}\) denotes the resulting node embedding. In a similar way, edge features are encoded to map raw attributes into a latent space.

\subsubsection{MPNN} The message function computes the information exchanged between nodes, factoring in the features of source nodes, target nodes, and connecting edges. This function is realized through the GatedGCN layer architecture \cite{Bresson2017arxiv}, which leverages information of the embeddings of previous layers. The MPNN layer uses the \textit{message-passing} framework as each node sends a message to its neighbors based on what it knows about itself, its neighbors, and how they're connected. The message function is expressed as:
\begin{equation} \label{eq:message}
    m_{ij}^{(l)} = M^{(l)}(h_i^{(l-1)}, h_j^{(l-1)}, e_{ij})
\end{equation}
where the message $m_{i\rightarrow j}^{(l)}$ at layer $l$ is determined by the embeddings of the features of the sending node $i$ and the receiving node $j$ from the previous layer, alongside the embeddings of the features from edge $e_{ij}$ connecting them.

The aggregation function is essential for summarizing the information from a node's neighbors.
It is defined as:
\begin{equation}
    \hat{m}_j^{(l)} = \sum_{i \in N(j)} m_{i\rightarrow j}^{(l)}
    \label{eq:aggregationfunction}
\end{equation}
with $\hat{m}_j^{(l)}$ representing the aggregated message at node $j$, encompassing the sum of all messages from its neighboring nodes $N(j)$ at layer $l$. This process captures the local network topology's influence, ensuring nodes receive a comprehensive aggregation of the messages from their neighbors. 

Finally, the update function, $\psi^{(l)}$, integrates these messages to refine each node's state:
\begin{align}
\mathbf{h}_i^{(l+1)} = \psi^{(l)} \left( \mathbf{h}_i^{(l)}, \mathbf{m}_j^{\hat{l}} \right)
\end{align}

In this formulation, $\psi^{(l)}$ represents the layer-specific update function, and $\mathbf{h}_i^{(l+1)}$ is the updated state of node $i$ where $m_{j}^{\hat{l}}$ is calculated through Eq.~\ref{eq:aggregationfunction}. By applying $\psi^{(l)}$, each node's new state is a function of its own features and the aggregated influence from its neighbors.

\subsubsection{Decoder} The final part of the model architecture is the decoder layer, which is responsible for predicting the traffic flow for each link within the transportation network. This prediction is made at the edge level, leveraging the comprehensive information encoded in the embeddings of the nodes, as well as the features of the edges themselves. To generate a flow prediction for a given link, the decoder uses inputs that are the embeddings of the source node (\(\mathbf{h}_i^{(L)}\)), i.e., the final layer's representation of the starting point of the link, and the destination node (\(\mathbf{h}_j^{(L)}\)), i.e., the final layer's representation of the endpoint of the link, along with the edge features (\(\mathbf{e}_{ij}\)), i.e., the attributes of the link itself. These components are concatenated to form a unified feature vector for each edge. The concatenated vector is then processed through a MLP to predict the flow. Mathematically, this process can be described as follows:

\begin{align}
\mathbf{z}_{ij} = \text{CONCAT}\left(\mathbf{h}_i^{(L)}, \mathbf{h}_j^{(L)}, \mathbf{e}_{ij}\right)
\end{align}
\vspace{-4mm}
\begin{align}
\hat{y}_{ij} = \text{MLP}(\mathbf{z}_{ij})
\end{align}

Where \(\mathbf{z}_{ij}\) represents the concatenated feature vector for the edge connecting nodes \(i\) and \(j\), and \(\hat{y}_{ij}\) denotes the predicted traffic flow on that link. The function \(\text{CONCAT}\) symbolizes the concatenation operation.

\subsubsection{Loss Function} \label{Loss Function}
The training of our model is directed by a L1 norm loss function that measures the deviation between the model's predicted traffic flows and the actual observed flows. The accuracy of the prediction is quantified by the mean absolute error (MAE) between predicted (\(\hat{y}_{ij}\)) and true traffic flow (\(y_{ij}\)) as:
\begin{align}
L_{\text{pred}} = \frac{1}{N} \sum_{ij} \left|\hat{y}_{ij} - y_{ij}\right|
\end{align}

with \(N\) representing the network's total link count.

\section{Experiments}

In this section, we detail the conducted experiments, describe the generation process of the dataset used for training and testing and discuss the obtained results.

\subsection{Dataset} \label{Dataset}

Our research employed the Sioux Falls network \citep{tnfr2023}, a mid-sized urban transport network consisting of 76 links and 24 nodes. In our dataset 11 of the nodes are defined as centroids meaning they have non-null demand in the OD matrix. For the remaining 13 nodes, the demand in the OD is set to 0. To train our GNN model effectively, we generated a synthetic dataset for the input parameters using Latin Hypercube Sampling (LHS) \citep{10.1145/167293.167637}. We configured 10,000 distinct samples, each maintaining the network topology but with unique demand and edge feature characteristics. Parameters such as Origin-Destination (OD) demand flow, free flow link speed, and capacity were set within realistic ranges (0 to 1,500 vehicles per OD pair, 45 to 80 km/hr, and 4,000 to 26,000 respectively) to ensure relevance to real-world scenarios. The application of the LHS allows us to define 10,000 scenarios that are meaningfully different between each other and maintains the overall levels of demand, capacity and speed limit across the network. PTV Visum \citep{ptvvisum} was used to simulate traffic flows for each scenario using the SUE assignment, serving as the ground truth for training our model.

\subsubsection{Data Preprocessing and Cleaning}
To enhance the model's training process and ensure that no single feature dominates others due to scale differences, we applied a MinMax normalizer to the input and output features. This normalization process transforms the feature values to a common scale, ranging from 0 to 1, thus preserving the original distribution of the features. Finally, after the preprocessing and cleaning phases were completed, the dataset was divided into training, validation, and test sets, with respective proportions of 60\%, 20\%, and 20\%. This division allows us to efficiently train our model, validate its performance, and finally, test its predictive performance on unseen data.

\subsubsection{Out-of-distribution data generation}

To assess the ability of our GNN model to generalize beyond the in-distribution (ID) conditions used during training, we constructed an out-of-distribution (OOD) dataset by systematically perturbing key network features, including link capacities, speed limits, and Origin-Destination (OD) demand. The OOD data generation process involved selecting a subset of links or nodes from each traffic network scenario and introducing controlled changes to the features. Specifically, for each feature (OD, capacity, and speed limit), between 10\% and 90\% of the network links or nodes were randomly chosen for modification. For the selected links or nodes the feature values were altered by up to 25\% beyond the original range defined for ID data. These modifications were applied to 50 randomly selected scenarios from the original 10,000 ID dataset for each perturbation level. After modifying the features, the traffic flows were recalculated using the Stochastic User Equilibrium (SUE) assignment in PTV Visum. 

\subsection{Training and Evaluation}
Our models were developed in Python, utilizing the \textit{PyTorch} \citep{DBLP:journals/corr/abs-1912-01703} and \textit{Pytorch Geometric} \citep{fey2019fast} libraries. The models have been built by exploiting the code from \cite{rampasek2022GPS}.
The experiments were conducted on a Linux server equipped with 128 GB RAM, 2x NVIDIA GeForce RTX 2080 TI GPU, and a Threadripper 2950X (40M Cache, 3.4 GHz base). The proposed MPNN has 64 hidden units in each of the 6 GatedGCN layers—reflecting the diameter of the Sioux Falls network (which ensures that information is exchanged between the farthest nodes in the network). Non-linearity between the layers is introduced using the ReLU activation function. During the training phase, we aimed to minimize the error in the L1 norm. The AdamW optimizer was chosen with a learning rate set to 0.001. The model was trained over 200 epochs in batches of 32 samples. To counteract overfitting, we implemented an early stopping criterion. The proposed model is benchmarked against three baseline models: the first is an MLP with access to all graph features and the graph structure, the second is an MPNN using the GCNConv layer which only uses the node features, and the third a mean comparison that provides the mean edge flow for each edge across all simulations. 

The MLP baseline comprises five hidden layers, each with 512 neurons, and uses ReLU activation functions after each layer. The parameters of the MLP were tuned on validation data, and the MSE loss was used as the loss function. The model was trained on flattened tensors containing all node features, edge features, and the graph's structure, represented by its flattened adjacency and incidence matrices. We highlight that providing the MLP with the flattened adjacency matrix and incidence matrix allow us to make a fair test of whether the architectural difference in how MPNNs use the adjancency in processing the edge and node features lead to increased performance. 
The GCNConv baseline was implemented using the GraphGPS framework from \citep{rampasek2022GPS}. This baseline utilized the GCNConv layer instead of the GatedGCN used in our proposed MPNN model. Unlike the proposed GatedGCN, the GCNConv baseline only leverages node features and lacked access to edge-specific information such as capacity, free-flow travel time, and speed limits. 
Finally, the mean comparison served as a simple yet informative baseline, providing the mean edge flow across all test samples, independent of input features. Notably, we compute the mean using the test data and later the OOD data, rather than the training data. This deliberate choice ensures a fairer and more transparent assessment of model performance relative to the true distribution of the target variable. By using mean of the test data, we set a more challenging benchmark, as models must outperform the actual mean flow observed in the test set rather than an estimate from the training data.

\subsection{Results for in-distribution performance}
In this section, we evaluate the performance of learning-based models as metamodels for equilibrium flow simulations. Our goal is to determine whether any of the models can achieve error rates low enough for practical metamodeling of traffic flows, with a specific focus on whether the proposed MPNN outperforms baseline models. These comparisons align with our broader objective of assessing whether learning-based approaches can serve as a viable alternative to traditional traffic simulators, offering prediction accuracy suitable for both real-time and large-scale metamodeling applications. All models were trained for 200 epochs with early stopping based on validation error. Their performance was subsequently evaluated on a hold-out test set.  Table~\ref{tab:performance_metrics} shows that the GatedGCN model (in bold) achieves better performance across all metrics.

\begin{table}[]
\centering
\resizebox{0.95\columnwidth}{!}{%
\begin{tabular}{lcccc}
\toprule
\textbf{Model}      & \textbf{MAE}     & \textbf{R\(^2\)}    & \textbf{MSE}     & \textbf{RMSE}    \\ 
\midrule
\textbf{GatedGCN}    & \textbf{0.02899}& \textbf{0.95210}& \textbf{0.00154}& \textbf{0.03921}\\ 
MLP                 & 0.03077          & 0.94808          & 0.00167          & 0.04082          \\ 
GCN                 & 0.05931& 0.80610& 0.00622& 0.07889\\ 
Mean                & 0.06565          & 0.77133          & 0.00734          & 0.08567          \\ 
\bottomrule
\end{tabular}
}
\captionsetup{font=small, width=0.95\columnwidth} 
\caption{Performance evaluation of models on test data across multiple metrics, for in-distribution predictions. The GatedGCN outperforms the other models across all metrics.} 
\label{tab:performance_metrics}
\vspace{-4mm}
\end{table}

\begin{figure}[]
    \centering
    \includegraphics[width=0.95\columnwidth]{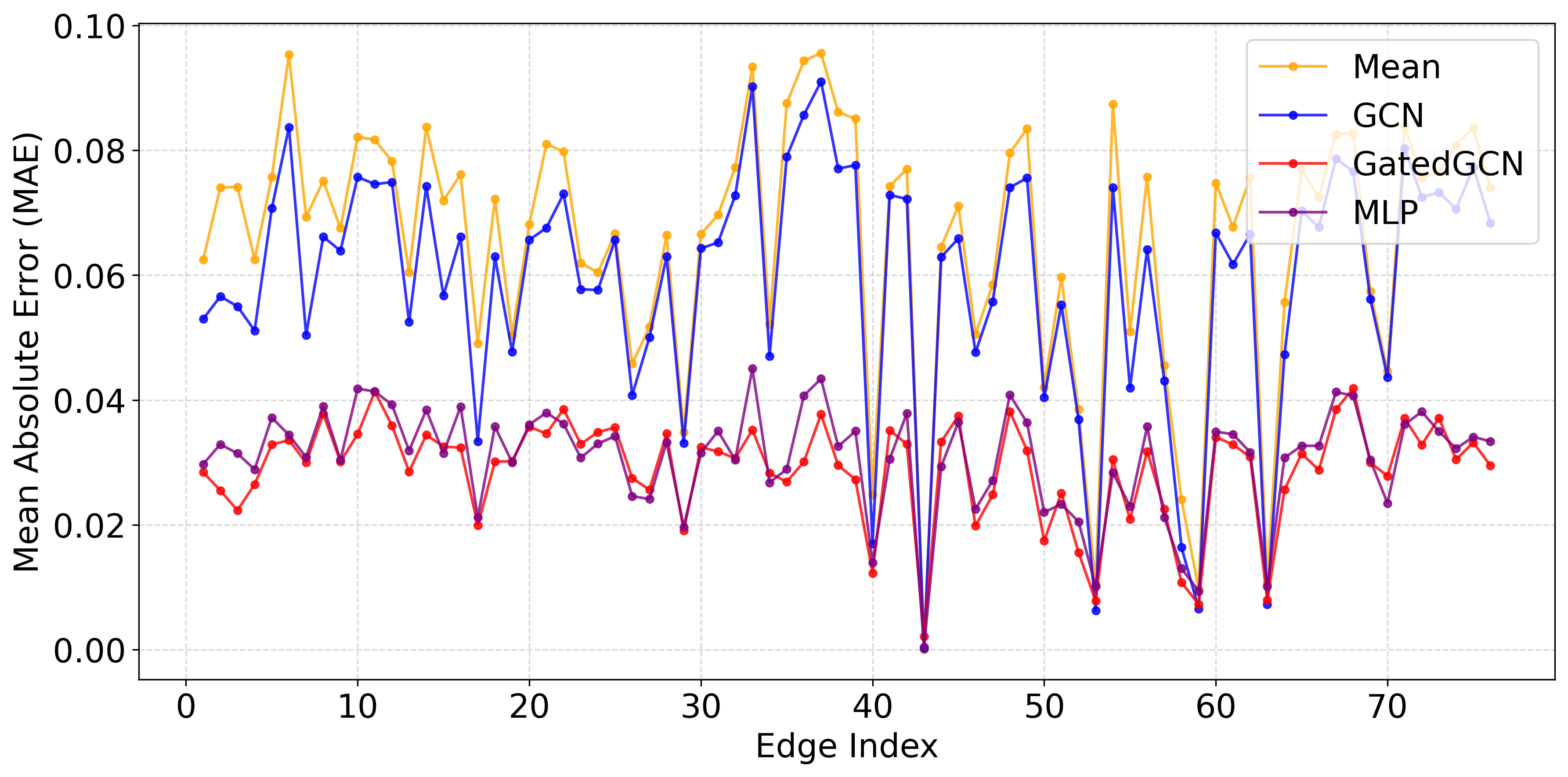}
    \captionsetup{font=small, width=0.95\columnwidth} 
    \caption{Mean absolute error (MAE) of all models per edge. The performance of the MLP and GatedGCN is best across all edges. Noticeably, some edges have lower MAE where the GCN and mean comparison }\label{fig:MAE1}
    \vspace{-6.5mm}
\end{figure}

In comparing the GatedGCN with the MLP, our study highlights how the MLP performs surprisingly well on this dataset. Indeed, as it will be showed in the following, when the node features become predominant on the link flows, the MLP may even perform better than other methods. When comparing the GatedGCN and MLP to the GCNConv, a critical distinction is the GatedGCN's ability to utilize edge features. The GCNConv, while effective in certain contexts, does not inherently process edge features, which leads to suboptimal performance as the edge characteristics are clearly crucial for predicting flows. 
Finally, the mean comparison in Table~\ref{tab:performance_metrics} obtains the worst performance, and hence, all models (including the GCN that only has access to node features) learn to use the graph features to predict edge flows. An overview of the MAE for each edge is provided in Fig.~\ref{fig:MAE1} which again shows that a traditional MLP has a similar performance in-distribution as the proposed GatedGCN.
One of the explanations for this is that we provided the MLP with access to the same information as the GatedGCN (including the adjacency matrix and incidence matrix). The GatedGCN should theoretically be able to better process this information through the structural relationship given to the GatedGCN by distributing the features onto the nodes and edges of the graph. However, one of the limitations for the GatedGCN is that nodes that are non-centroids have node feature vectors with only zeros. This might weaken the GatedGCNs ability to learn from the demand features and it has to rely more on edge features (as the weights on the non-centroid nodes vanishes when multiplied with zeros). This is not an issue for the MLP as the zone-only OD matrix  $(11 \times 11)$ is flattened directly into the feature vector and does not contain any zeros. 

Even though the results in-distribution do not show a clear advantage of the message-passing neural network compared to traditional neural networks, the benefits of the MPNN's arise when assessing the models' performance out-of-distribution, specifically when edge features become predominant (e.g., when capacities are OOD). Generally, the results show great potential for using learning-based models to approximate equilibrium flows providing a great alternative to computationally expensive iterative simulation algorithms. 

\subsection{Experiments on Out-of-Distribution Dataset}

This section presents experiments to evaluate the models' robustness and generalization under out-of-distribution (OOD) conditions, where input features (OD demand, capacity, speed limit) deviate from the training data distribution.

\subsubsection{Changes to speed limit}
\begin{figure}[]
    \centering
    \includegraphics[width=0.95\columnwidth]{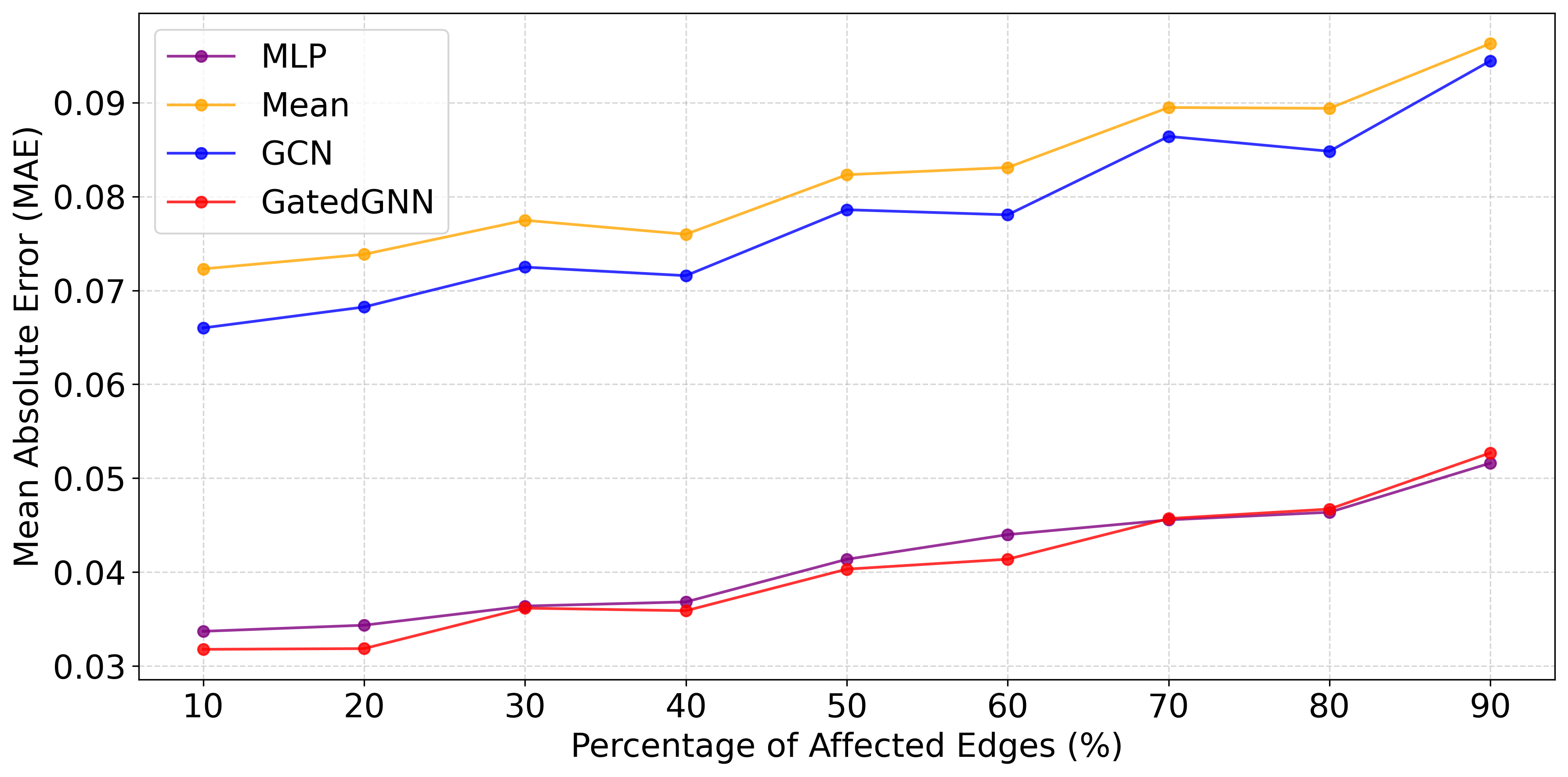}
    \captionsetup{font=small, width=0.95\columnwidth} 
    \caption{Mean absolute error across all models when changing x\% of the links to up to 25\% out-of-distribution speed limits.}
    \label{fig:OOD_speed}
    \vspace{-2.5mm}
\end{figure}

In Fig.~\ref{fig:OOD_speed} we see a similar performance ranking as for the in-distribution results. The MLP is performing surprisingly well out-of-distribution even with changes in the speed limits for 90\% of the edges. The MLP and the GatedGCN outperform the GCNConv and mean comparison with the same absolute distance in MAE as the percentage of edge features changed is increasing.

\subsubsection{Changes to capacities} 
\begin{figure}[]
    \centering
    \includegraphics[width=0.95\columnwidth]{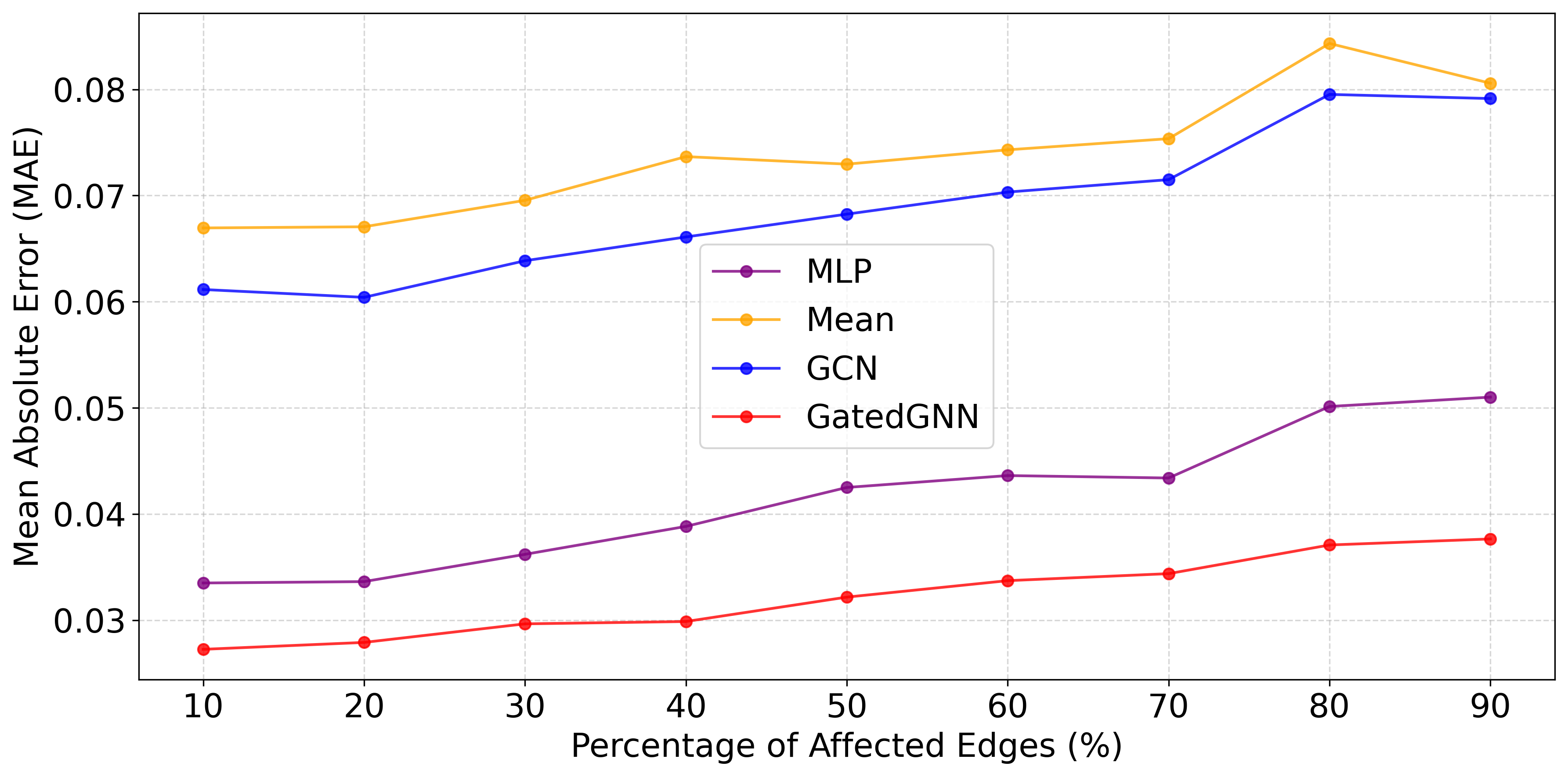}
    \captionsetup{font=small, width=0.95\columnwidth} 
    \caption{Mean absolute error across all models when changing x\% of the links to up to 25\% out-of-distribution capacities.}
    \label{fig:OOD_capacities}
    \vspace{-6.5mm}
\end{figure}

In Fig.~\ref{fig:OOD_capacities}, we observe that the GatedGCN outperforms the other baseline models, suggesting that it has more effectively captured the underlying impact of capacity changes on edge flow. Additionally, it is noteworthy that the increase in MAE remains within reasonable limits ($<0.4$ MAE), even when 90\% of the links have capacities with out-of-distribution values.

\subsubsection{Changes to OD matrix} 
\begin{figure}[]
    \centering
    \includegraphics[width=0.95\columnwidth]{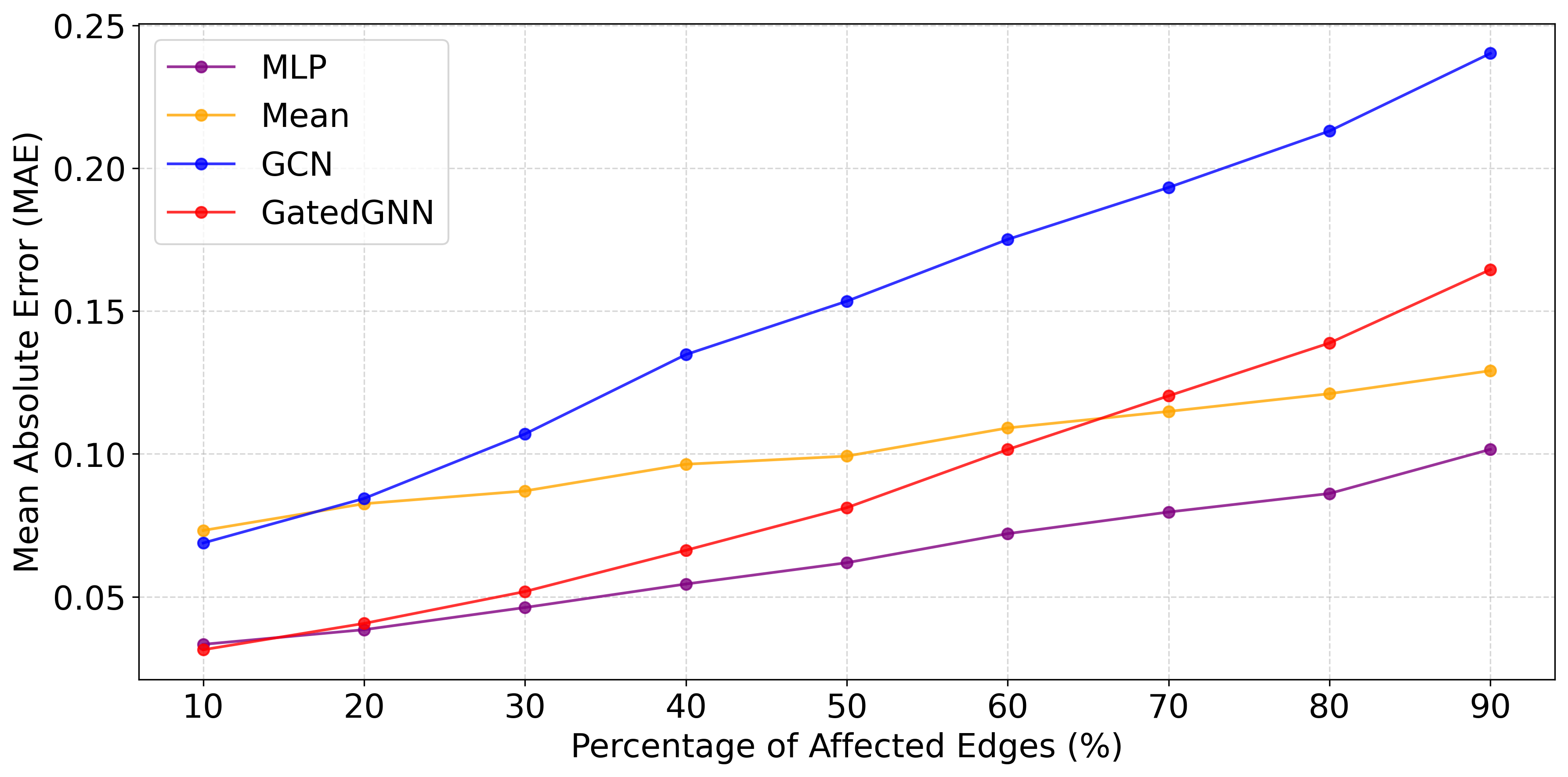}
    \captionsetup{font=small, width=0.95\columnwidth} 
    \caption{Mean absolute error across all models when changing x\% of the nodes to up to 25\% out-of-distribution demand flows.}
    \label{fig:OOD_demand}
    \vspace{-4mm}
\end{figure}

In Fig.~\ref{fig:OOD_demand}, we observe that both the MLP and GatedGCN maintain stable performance when 10–20\% of OD pairs experience increased demand. However, as the proportion of OD pairs with increased demand reaches 70\%, the GatedGCN starts to underperform relative to the mean comparison, while the MLP consistently remains the best-performing model from as early as 20\% OD pair increase.

The overall failure of all models under increased OD demand likely stems from the limited variation in total demand during training, a consequence of the Latin Hypercube sampling strategy. In contrast, the OOD data introduces up to a 25\% increase in total network flow, creating a significant distribution shift. The limited variation in total demand during training may have led the models to focus more on redistributing a fixed total demand across links rather than accurately capturing the absolute scale of OD values. The MLP achieves the best performance, likely due to its direct access to demand information through the flattened OD matrix. In contrast, the GatedGCN may suffer from signal attenuation across layers and a reduced reliance on node features, exacerbated by the prevalence of zero feature vectors. This suggests that the MLP leverages OD features more explicitly for flow prediction, while the GatedGCN relies more heavily on edge features.

\section{Limitations and future research directions}
Future work may strive to address some of the limitations of the current paper. A first limitation of the presented work lies in the data generation process described in Section~\ref{Dataset}. The use of the Latin Hypercube on the single parameters of each simulation leads to a training dataset with different permutations of OD matrices, capacities and speed limits, but with a similar uniform distributions across nodes and edges. Addressing this limitation can increase the variability of link flows that can help the models better learn the underlying relationship between input data, the graph, and the equilibrium algorithm enhancing the performance out-of-distribution. In addition, the performance of the models out-of-distribution should be assessed also for scenarios in which capacity, speed limits, and demand are OOD simultaneously, to test the models' performance during extreme scenarios. 
Even though the performance of the models were good, we believe that adding additional graph-based features (that can inform about flow or directionality) can enhance the performance of the models in and out-of-distribution. Also, we believe that custom MPNN architectures designed to better mimic the algorithmic structure can further enhance performance.
Besides, we hypothesize that the power of the MPNNs would be better exploited in more complex networks where the graph structure of the problem may have a bigger impact, which would in turn make it more challenging for the MLP to perform as well. For example, larger networks could see widely different routing choices arising from changes in the inputs. 

\section{Conclusion}

This study introduced a novel data-driven approach for traffic assignment utilizing a message-passing graph neural network. We found its performance on flow prediction to be satisfactory; however, traditional ML architectures with access to the same information (such as an MLP) perform comparatively well. The MPNN successfully incorporates edge features into the learning process and achieves high performance out-of-distribution on changing capacities and speed limits. This highlights the key advantage of GNNs, namely, leveraging the edge feature to capturing spatial dependencies. Moreover, our results underline the potential of learning-based models, in particular GNNs, to approximate traffic flow efficiently, enabling real-time decision-making without the need for computationally expensive simulations.

\section*{Acknowledgments}
The work presented in this article is supported by Novo Nordisk Foundation grant NNF23OC0085356.

\input{main.bbl}

\end{document}